\documentclass{article}

\usepackage{microtype}
\usepackage{graphicx}
\usepackage{subfigure}
\usepackage{booktabs} % for professional tables
\usepackage{enumerate}
\usepackage{enumitem}
\usepackage{multirow,bm}
\usepackage{bbm}
\usepackage{amssymb}
\usepackage{xcolor}

\usepackage{hyperref}

\usepackage[accepted]{icml2023}

\usepackage{amsmath}
\usepackage{amssymb}
\usepackage{mathtools}
\usepackage{amsthm}

\usepackage[capitalize,noabbrev]{cleveref}

\theoremstyle{plain}
\newtheorem{theorem}{Theorem}[section]
\newtheorem{proposition}[theorem]{Proposition}

\theoremstyle{definition}

\theoremstyle{remark}

\usepackage[textsize=tiny]{todonotes}

\icmltitlerunning{\method{}: A Coupled Contrastive Framework for Unsupervised Domain Adaptive Graph Classification}
\def\method{CoCo}

\begin{document}

\twocolumn[
\icmltitle{\method{}: A Coupled Contrastive Framework for \\ Unsupervised Domain Adaptive Graph Classification}



\icmlsetsymbol{equal}{*}

\begin{icmlauthorlist}
\icmlauthor{Nan Yin}{a}
\icmlauthor{Li Shen}{b}
\icmlauthor{Mengzhu Wang}{c}
\icmlauthor{Long Lan}{c}
\icmlauthor{Zeyu Ma}{e}
\icmlauthor{Chong Chen}{f}
\icmlauthor{Xian-Sheng Hua}{f}
\icmlauthor{Xiao Luo}{g}
\end{icmlauthorlist}

\icmlaffiliation{a}{Department of Machine Learning, Mohamed bin Zayed University of Artificial Intelligence, Abu Dhabi, UAE}
\icmlaffiliation{b}{JD Explore Academy, Beijing, China}
\icmlaffiliation{c}{Department of Computer Science and Technology, National University of Defense Technology, Changsha, China}
% \icmlaffiliation{d}{School of ZZZ, Institute of WWW, Location, Country}
\icmlaffiliation{e}{School of Computer Science and Technology, Harbin Institute of Technology, Shenzhen, China}
\icmlaffiliation{f}{Terminus Group, Beijing, China}
\icmlaffiliation{g}{Department of Computer Science, University of California, Los Angeles, USA}

\icmlcorrespondingauthor{Xiao Luo}{xiaoluo@cs.ucla.edu}

\icmlkeywords{Machine Learning, ICML}

\vskip 0.3in
]



\printAffiliationsAndNotice{} % otherwise use the standard text.

\begin{abstract}
Although graph neural networks (GNNs) have achieved impressive achievements in graph classification, they often need abundant task-specific labels, which could be extensively costly to acquire. A credible solution is to explore additional labeled graphs to enhance unsupervised learning on the target domain. However, how to apply GNNs to domain adaptation remains unsolved owing to the insufficient exploration of graph topology and the significant domain discrepancy.  
In this paper, we propose \underline{Co}upled \underline{Co}ntrastive Graph Representation Learning (\method{}), which extracts the topological information from coupled learning branches and reduces the domain discrepancy with coupled contrastive learning.
\method{} contains a graph convolutional network branch and a hierarchical graph kernel network branch, which explore graph topology in implicit and explicit manners. Besides, we incorporate coupled branches into a holistic multi-view contrastive learning framework, 
which not only incorporates graph representations learned from complementary views for enhanced understanding, but also encourages the similarity between cross-domain example pairs with the same semantics for domain alignment.
Extensive experiments on popular datasets show that our \method{} outperforms these competing baselines in different settings generally.
\end{abstract}

% abstract还需要update一下，整体来说motivation并不是很清楚。按照一下的句子来展开
% 第一句：介绍整个setting的背景，以及这个setting有什么挑战（background）
% 第二句：现有的方法在解决这类setting的时候，有什么缺点以及局限和挑战 （motivation）
% 第三句：针对以上的这些问题，我们提出了COCO这个算法，他利用了什么工具来解决这个问题（highlevel的介绍我们的方法）
% 第四句：具体来介绍 COCO 的设计范式和工作原理，由什么模块组成，每个模块的作用和功能，以及这些模块分别解决了什么问题，体现出technical contribution（how and why ， compared with existing work），这句话如果比较长的话，可以split一下
% 第五句： 我们在XXX实验上，和XXX sota对比，取得了什么提升，说明了我们方法的有效性。

% 同样的， intro本质上就是 abstract的展开
% 第一段： abstract的第一句的展开
% 第二段： abstrac 的第二句的展开
% 第三段： abstract的第三句和第四句的展开（说清楚我们的方法，以及与existing work的不同）
% 第四段： abstract的第五句展开，说清楚我们做了什么实验，与什么sota进行了对比，有了多少的 acc gains
% 最后，summarize our contributio 
\section{Introduction}

Recently, graph-structured data has flourished in a number of fields including chemistry and bioinformatics. 
Among various graph-based machine learning tasks, graph classification seeks to predict the properties of whole graphs~\cite{wang2021mixup,kong2022robust}, and a
variety of machine learning algorithms have been put forward for the problem~\cite{yoo2022model,feng2022kergnns,zhang2021deep,yang2022dual}. These methods mostly fall under the category of graph neural networks (GNNs). Following the paradigm of message passing~\cite{kipf2017semi}, GNNs learn representations in the graph by stacking multiple neural network layers, each of which transfers semantic information from topological neighbors to centroid nodes. A readout function eventually combines all of the node representations into a graph-level representation. In this way, GNNs are capable of integrating graph structural information into graph representations implicitly, thus facilitating downstream graph classification.

In spite of their certain progress, modern GNN algorithms are typically trained under supervision, necessitating a large quantity of labeled data~\cite{kipf2017semi,xu2019powerful,bodnar2021weisfeiler,BaekKH21}. However, label annotation in the graph domain is either prohibitively expensive or even impossible to acquire~\cite{xu2021infogcl,suresh2021adversarial}. For instance, ascertaining the pharmacological effect of drug molecule graphs involve costly experiments on living animals. Due to the scarcity of labeled annotations, the bulk of current algorithms performs poorly in practice. To solve this issue, we observe that there are often a substantial number of graph samples from a different but relevant domain and their labels are easily available. In this spirit, this work studies the problem of unsupervised domain adaptive graph classification, a practical task to predict graph properties with both labeled source graphs and unlabeled target graphs. 

However, designing an effective domain adaptive graph classification framework is non-trivial due to the following major challenges. (1) \textit{How to sufficiently extract topological information under the scarcity of labeled data?} Recent GNN algorithms~\cite{xu2019powerful} are trained in an end-to-end manner following the paradigm of message passing, which merely extracts structural information in an implicit manner. Learning implicit topological knowledge is not sufficient under the shortage of supervised signals in target domains. Although a range of graph kernels~\cite{borgwardt2005shortest,shervashidze2011weisfeiler} have been put forward to explicitly extract structural signals, they are usually derived in an unsupervised manner, failing to extract related information with supervised information. (2) \textit{How to effectively reduce the domain discrepancy in the graph space?} Different from the node classification problem, in this scenario, we face a variety of graphs in a complicated space instead of a single graph.
The potential domain discrepancy exacerbates the hardness of effective representation learning for graph samples. As a consequence, although a range of domain adaption methods has been proposed in computer vision~\cite{he2022secret,huang2022category,nguyen2021kl,xiao2021dynamic}, they cannot be directly employed to learn domain-invariant and discriminative graph-level representations for our problem.

To tackle these challenges, we propose a holistic method named \underline{Co}upled \underline{Co}ntrastive Graph Representation Learning (\method{}) for unsupervised domain adaptive graph classification. 
To holistically extract topological information, \method{} incorporates coupled branches, which learn structural knowledge using end-to-end training from both implicit and explicit manners, respectively. On the one hand, a GCN branch leverages the message-passing paradigm to implicitly extract graph topological knowledge. On the other hand, a hierarchical GKN branch leverages a graph kernel~\cite{shervashidze2009efficient} to compare samples with learnable filters to explicitly incorporate topological information into graph representations. To achieve effective domain adaptation, we integrate topological information from two branches into a unified multi-level contrastive learning framework, which contains cross-branch contrastive learning and cross-domain contrastive learning. To couple the structural information from two complementary views, cross-branch contrastive learning seeks to promote the agreement of two branches for each graph sample, therefore producing high-quality graph representations with comprehensive semantics. To reduce the domain discrepancy, we first calculate the pseudo-labels of target data in a non-parametric manner and then introduce cross-domain contrastive learning, which minimizes the distances between cross-domain sample pairs with the same semantics compared to those with different semantics. More importantly, we theoretically demonstrate that our cross-domain contrastive learning can be formalized as a problem of maximizing the log-likelihood solved by Expectation Maximization (EM). 
Extensive experimented conducted on various widely recognized benchmark datasets for graph classification reveal that the proposed \method{} beats a range of competing baselines by a considerable margin. 

The main contributions can be summarized as follows:
\begin{itemize}[itemsep=2pt,topsep=0pt,parsep=0pt]
\item We introduce a new approach for unsupervised domain adaptive graph classification, named \method{}, which contains a graph convolutional network branch and a hierarchical graph kernel network branch to mine topological information from different perspectives.
\item On the one hand, cross-branch contrastive learning encourages the agreement of coupled modules to generate comprehensive graph representations. On the other hand, cross-domain contrastive learning reduces the distances between cross-domain pairs with the same semantics for effective domain alignment.
\item Comprehensive experiments on various widely-used graph classification benchmark datasets demonstrate the effectiveness of the proposed \method{}.
\end{itemize}

\section{Related Work}

\subsection{Graph Classification}
Graph classification has been a long-standing problem with various applications in social analysis~\cite{fan2019graph,song2016influential,liao2021learning,wu2020comprehensive,ju2023comprehensive,ju2022glcc} and molecular property prediction~\cite{hassani2022cross,yin2022autogcl}. Early efforts to this problem almost turn to graph kernels including Weisfeiler-Lehman kernel~\cite{shervashidze2011weisfeiler} and random walk kernel~\cite{kang2012fast}, which can identity graph substructures through graph decomposition. However, these methods cannot be well applied to large-scale graphs due to high computational costs. To tackle this, a variety of graph neural networks have been put forward in recent years~\citep{kipf2017semi,xu2019powerful,bodnar2021weisfeiler,BaekKH21,ju2023tgnn}. Typically, these algorithms obey the message passing paradigm to iteratively update node representations, followed by a graph pooling function that generates graph-level representations for downstream classification. Hence, these methods usually explore topological information merely in an implicit way. However, recent research has demonstrated that this paradigm is inadequate for detecting structural motifs in graphs such as rings~\cite{long2021theoretically,chen2020convolutional,cosmo2021graph}.
To tackle this issue, besides implicitly exploring graph topological knowledge using the graph convolutional network branch, \method{} utilizes a hierarchical graph kernel network to explicitly explore graph topology, which enhances the classification performance in a domain adaptive framework.

\begin{figure*}[t]
  \centering
  \includegraphics[scale=1.05]{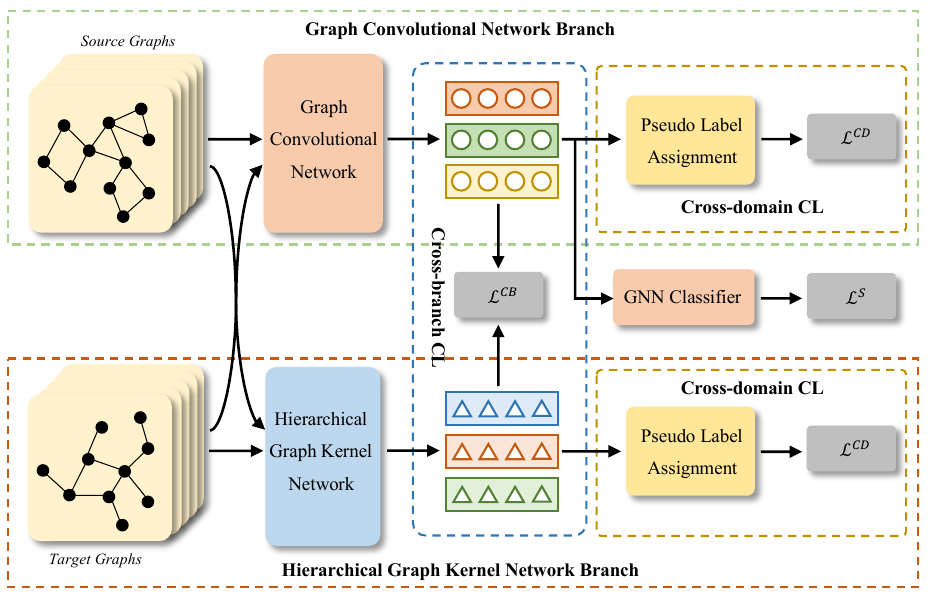}
  \caption{An overview of the proposed \method{}. Our \method{} feeds source graphs and target graphs into two branches (i.e., GCN branch and HGKN branch). Cross-branch contrastive learning compares graph representations from two branches to achieve while cross-branch contrastive learning compares graph representations from two domains with the guidance of pseudo-labels.}
  \label{fig1}
\end{figure*}

\subsection{Unsupervised Domain Adaptation}
The purpose of unsupervised domain adaptation (UDA) is to transfer a model from a label-rich source domain to a label-scarce target domain~\cite{he2022secret,huang2022category,nguyen2021kl,wang2022cross}. This problem is substantially investigated in the field of computer vision with applications in image classification~\cite{mirza2022dua}, image retrieval~\cite{zheng2021group} and semantic segmentation~\cite{kundu2022amplitude}. In general, domain alignment is the key to successful domain adaptation. The early attempts usually explicitly reduce domain discrepancy using statistical metrics including maximum mean discrepancy~\cite{long2015learning} and Wasserstein distance~\cite{shen2018wasserstein}. Recently, adversarial learning-based approaches are the predominant paradigm for UDA~\cite{long2018conditional}. These methods usually employ a gradient reversal layer~\cite{ganin2016domain} to render deep features resistant to domain shifts. In addition, discrimination learning under label scarcity is an important challenge for UDA~\cite{xiao2021dynamic,liang2020we} and pseudo-labeling techniques are popular strategies~\cite{lee2013pseudo,assran2021semi} to tackle this problem in the target domain. Despite the significant advance of UDA in computer vision, it is still underexplored in the graph domain. 
In this work, we study the emerging and practical problem of unsupervised domain adaptive graph classification, which employs labeled source graphs to improve the classification performance on unlabeled target graphs. Different from existing domain adaptation methods~\cite{yin2022deal,mirza2022dua,wei2021metaalign}, our \method{} explores semantics information from different views and conduct coupled contrastive learning for effective domain adaptation.

\section{Methodology}

The overview of the proposed \method{} framework for unsupervised domain adaptive graph classification is illustrated in Figure \ref{fig1}. The core of our \method{} is to provide two complementary views to explore graph topology using coupled branches. From an implicit view, we utilize a graph convolutional network branch to infer topological information (see Section \ref{GCN_branch}), while a hierarchical graph kernel network branch is adopted to compare graph samples with learnable filters (see Section \ref{HGKN_branch}), which explicitly summarizes topological information. Moreover, we incorporate the two branches into a holistic multi-view contrastive learning framework (see Section \ref{MCL_branch}). On the one hand, we perform cross-branch contrastive learning to encourage the agreement of two branches for representations containing comprehensive structural information. On the other hand, cross-domain contrastive learning aims to minimize the distances between cross-domain sample pairs with the same semantics compared to those with different semantics. 

\subsection{Problem Formulation}
Given a graph $G = (\mathcal{V}, \mathcal{E})$, in which $\mathcal{V}$ is the set of nodes and $\mathcal{E} \subseteq \mathcal{V} \times \mathcal{V}$ denotes the set of edges. There is also a node feature matrix $\bm{X} \in \mathbb{R}^{|\mathcal{V}| \times d}$, where each row $\bm{x}_v \in \mathbb{R}^{d}$ denotes the feature of node $v \in \mathcal{V}$, $|\mathcal{V}|$ is the number of nodes, and $d$ denotes the dimension of node features.
In our problem, we have access to a labeled source domain $\mathcal{D}^s = \{(G_i^s, y_i^s)\}_{i=1}^{n_s}$ with $n_s$ samples and an unlabeled target domain $\mathcal{D}^t = \{G_j^t\}_{j=1}^{n_t}$ with $n_t$ samples. $\mathcal{D}^s$ and $\mathcal{D}^t$ share the label space, i.e., $\mathcal{Y} =\{1,2,\cdots, C\}$ with different distributions in the data space. We expect to train the graph classification model using both $\mathcal{D}^s$ and $\mathcal{D}^t$, and attain high accuracy on the test dataset on the target domain.

\subsection{Graph Convolutional Network Branch}\label{GCN_branch}

In this branch, we utilize a graph convolutional network to implicitly extract topological information.
Graph convolutional networks (GCNs) typically follow the message-passing scheme to embed the structural and attribute information into node representations, which have shown their superior capability in graph classification~\cite{gilmer2017neural,kipf2017semi,xu2019powerful}. 
In detail, for each node, we start by aggregating the embedding vectors of all its neighbors at the previous layer. Then the node representation is updated iteratively by fusing the representation from the last layer with the aggregated neighbor embedding. In formulation, the representation of node $v\in G$ at the $l$-th layer $\bm{h}_v^{(l)}$ is calculated as follows:
\begin{equation}
\bm{h}_{v}^{(l)}= \operatorname{COM}^{(l)}_{\theta}\left(\bm{h}_{v}^{(l-1)}, \operatorname{AGG}^{(l)}_{\theta} \left(\left\{\bm{h}_{u}^{(l-1)}\right\}_{u \in \mathcal{N}(v)}\right) \right), \nonumber
\end{equation}
where $N(v)$ denotes the neighbors of $v$. $\operatorname{AGG}^{(l)}_{\theta}$ and $\operatorname{COM}^{(k)}_{\theta}$ denote the aggregation and combination operations parameterized by $\theta$ at the $l$-th layer, respectively. Ultimately, we adopt an extra $\operatorname{READOUT}$ function to summarize the node representations at the last layer into a graph-level representation. In formulation,
\begin{equation}
g_{\theta}\left(G \right)=\operatorname{READOUT}\left(\left\{\bm{h}_{v}^{(L)}\right\}_{v \in V}\right),
\end{equation}
where $g_{\theta}\left(G\right)$ denotes the graph-level representation and the network parameters are denoted by $\theta$. 

\begin{figure}[t]
  \centering
  \includegraphics[scale=0.65]{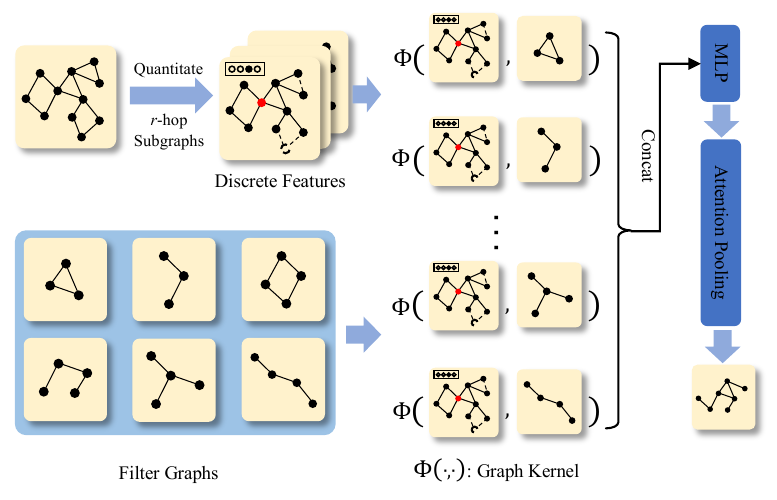}
  \caption{An illustration of the hierarchical graph kernel network branch. HGKN compares each $r$-hop subgraph with filter graphs to generate node features, followed by an attention-based pooling operation to coarsen the graph.}
  \label{GKN}
\end{figure}

\subsection{Hierarchical Graph Kernel Network Branch}\label{HGKN_branch}

However, GCNs are insufficient in collecting sophisticated high-order structural information such as rings~\cite{long2021theoretically,cosmo2021graph}. Consequently, it is anticipated to explore graph topology in an explicit manner as a compliment.
To accomplish this, we introduce a hierarchical graph kernel network (HGKN) based on a graph kernel (e.g., WL kernel, random walk kernel). Generally, at each hierarchical layer, it compares each $r$-hop subgraph with learnable filter graphs via the graph kernel to update node representations, followed by an attention-based pooling operation to coarsen the graph. After multiple hierarchical layers, a graph-level representation can be obtained in an end-to-end manner.

To be specific, for each graph, we first extract the local information of each node using its $r$-hop subgraph, which comprises all nodes reached by the central node within $r$ edges, along with all the edges between these selected nodes. To explore topological information from these subgraphs, we generate $M$ undirected learnable graphs with varying sizes serving as filters, i.e., $\{\tilde{G}_1^{(k)}, \cdots, \tilde{G}_M^{(k)} \}$ at the $k$-th layer. Each of the filter graphs $\tilde{G}_m^{(k)}$ is with a trainable adjacency matrix and each node is with an attribute. We expect these learnable filters to extract high-order structural information for better graph classification. 
However, most of the graph kernels usually accompany with discrete node attributes~\cite{cosmo2021graph,schulz2019necessity}. To tackle this, we introduce a quantization operation $Q(\cdot)$, which discretizes node attributes using clustering during network forwarding. Through $Q(\cdot)$, the continuous attributes are replaced by the discrete cluster assignments. Then, we compare the discrete input graph and the filter graphs using the graph kernel. 
\begin{equation}
    e_{v}^{(k)}(m) = \Phi (Q(S_v^{(k-1)}),\tilde{G}_m^{(k)}),
\end{equation}
where $S_v^{(k-1)}$ denotes the subgraph with center $v$ at the previous layer and $\Phi(\cdot,\cdot)$ denotes the given graph kernel. Then, we can update the node representation $\bm{e}_v^{(k)}\in\mathbb{R}^M$ by concatenating all the output of filters as follows:
\begin{equation}
    \bm{e}_v^{(k)} = [e_{v}^{(k)}(1), \cdots, e_{M}^{(v)}(M)].
\end{equation}
Finally, we utilize a multi-layer perceptron (MLP) $\psi^k(\cdot)$ to project the concatenated kernel values into node representations at the $k$-th layer, i.e., $\bm{x}_v^{(k)}=\psi^k(\bm{e}_v^{(k)})$.

To address the issue that the quantization operation cannot be compatible with stochastic gradient descent, we merely calculate the gradient with respect to the filters in each layer of the graph kernel network during backpropagation. This strategy has been extensively used when optimizing the deep hashing networks, exhibiting satisfactory performance empirically~\cite{tan2020learning,qiu2021unsupervised}. In addition, we utilize a discrete algorithm to update filter graphs using the edit operation which includes adding (removing) edges and modifying node attributes. To be specific, for each $\tilde{G}_m^{k}$, we conduct a random edit operation candidate $\tilde{G'}_m^{k}$ and determine whether to accept it based on the gradient of the loss with respect to the filters. In formulation, 
\begin{equation}
    \frac{\partial \mathcal{L}}{\partial \tilde{G}_m^{k}} = \sum_{v\in G}\frac{\partial \mathcal{L}}{\partial e_{v}^{(k)}(m) }\frac{\partial e_{v}^{(k)}(m)}{\partial \tilde{G}_m^{k}}.
\end{equation}
However, we cannot acquire ${\partial e_{v}^{(k)}(m)}/{\partial \tilde{G}_m^{k}}$. Instead, we utilize the difference between kernel values of candidate $\tilde{G'}_m^{k}$ and original filter $\tilde{G}_m^{k}$, i.e., $\Phi (Q(S_v^{(k-1)}),\tilde{G'}_m^{(k)})-\Phi (Q(S_v^{(k-1)}),\tilde{G}_m^{(k)}) $ to replace it. After estimating ${\partial \mathcal{L}}/{\partial \tilde{G}_m^{k}}$, we will accept the candidate if the gradient is above zero and vice versa.

As in convolutional neural network (CNN) architectures~\cite{passalis2018training}, after computing the graph kernel, we additionally introduce a pooling operation~\cite{lee2019self} using the attention mechanism to provide a hierarchical view while reducing the computational cost. 
To be specific, we utilize an MLP to produce score vector $Z\in \mathbb{R}^{|V^{(k)}|\times 1}$ for each graph, where $V^{(k)}$ represents the set of nodes at the $k$-th layer. On the basis of scores, the top $\lceil \rho |V^{(k)}|\rceil$ nodes are kept with an index set $idx$ where $\rho\in(0,1)$ is the pooling ratio. 
In the end, we calculate the pooled graph $G_{pool}^{(k)}$ with the attribute matrix $\bm{X}_{pool}^{(k)}$ and adjacent matrix $A_{pool}^{(k)}$. The learnt node representations are first concatenated into a matrix $\bm{X}^{(k)}$.
Let $\odot$ denote the broadcasted Hadamard product, and we have:
\begin{equation}
X^{(k)}_{pool}=X^{(k)}_{idx,:} \odot Z_{idx}, A^{(k)}_{pool}=A^{(k)}_{idx, idx},
\end{equation}
where $X^{(k)}_{idx,:}$, $A^{(k)}_{idx, idx}$ denote the node-indexed feature matrix and the row, column-indexed adjacency matrix. 

After multiple hierarchical graph kernel layers followed by graph pooling operations, we feed all the node feature $\bm{e}_v^{(K)}$ into a fully-connected layer followed by a sum-pooling to produce a graph-level representation denoted as $f_\phi(\mathcal{G})$.

\subsection{Multi-view Contrastive Learning Framework}\label{MCL_branch}

To fully exploit graph topology information derived from two branches,
we formalize a multi-view contrastive learning framework, which contains both cross-branch and cross-domain contrastive learning for effective domain adaptation. 

\noindent\textbf{Cross-branch Contrastive Learning.}  
Considering that the model learns graph semantics from complementary views, we contrast the graph representations from two branches to exchange their knowledge mutually, which enhances discrimination learning on target data under label scarcity.

Specifically, for each a graph $G_i$ in a source batch $\mathcal{B}^s$ and a target batch $\mathcal{B}^t$, we produce the embeddings from coupled branches, i.e., $\bm{z}_i=g_\theta(G_i)$ and $\tilde{\bm{z}}_i=f_\phi(G_i)$. Then, we introduce the InfoNCE loss to enhance the consistency cross coupled branches. In formulation,
\begin{equation}
    \mathcal{L}^{CB} \!=\! \frac{1}{|\mathcal{B}^s|+|\mathcal{B}^t|}\sum_{G_i\in \mathcal{B}^s\cup \mathcal{B}^t} \!\!-\!\log\frac{exp(\bm{z}_i \ast \tilde{\bm{z}}_i/ \tau)}{\sum_{G_{i'}\in \mathcal{B}^t} exp({\bm{z}_{i}\ast \tilde{\bm{z}}_{i'}/ \tau})},
\end{equation}
where $\tau$ represents a temperature parameter set to $0.5$ as in previous works \cite{he2020momentum}. Through providing challenging views by mining topological information in complementary manners, our proposed \method{} is a special kind of graph contrastive learning~\cite{you2020graph,you2021graph,yin2022autogcl}, hence producing discriminative graph-level representation with sufficient topological information. 

\noindent\textbf{Cross-domain Contrastive Learning.} 
However, with a serious domain shift in the graph space, the graph representations could remain biased and unreliable for downstream classification. Intuitively, the representations of source (target) samples should be close to target (source) samples with the same semantics. To achieve this, we need to generate pseudo-labels of target data as a preliminary. In light of the fact that learning a classifier is suboptimal and biased owing to the paucity of labels, we generate pseudo-labels in a non-parametrical manner by comparing the similarities between target graphs and source graphs. On this basis, we conduct cross-domain contrastive learning which minimizes the distances between cross-domain example pairs with the same semantics compared to those with different semantics. 

Taking the GCN branch as an example, we utilize a non-parametrical classifier to generate the pseudo-label for each $G_j^t$. In formulation, we have:
\begin{equation}\label{eq:ass}
\hat{p}_j^t=\sum_{(G_i^s, y_i^s) \in \mathcal{B}^s}\left(\frac{ \zeta \left(\bm{z}_{j}, g_{\theta}(G_i^s)\right)}{\sum_{(G_i^s, y_i^s) \in \mathcal{B}^s} \zeta\left(\bm{z}_{j}, g_{\theta}(G_i^s)\right)}\right) \bm{y}_i^s,
\end{equation}
where $\bm{y}_i^s$ denotes the one-hot label embedding and $\zeta \left(\bm{z}_{j}, g_{\theta}(G_i^s)\right)= exp(\bm{z}_{j}\ast g_{\theta}(G_i^s)/ \tau) $ denotes the similarities between two vectors. In Eq. \ref{eq:ass}, we involve a batch of labeled source data in the classifier for efficiency. The pseudo-labels can be easily derived from $\hat{p}_j^t$, i.e., $\hat{y}^t_j = \arg max(\hat{p}_j^t)$.

Then, we pull close the representations with the same semantics across domains to minimize domain discrepancy. To achieve this, we treat the source samples with identical labels as positives for each target sample. In formulation, we set $\Pi(j) = \{i|y_i^s = \hat{y}^t_j, G_i^s\in \mathcal{B}^s\}$ to represent the index of all positives in the mini-batch and the cross-domain contrastive learning objective is written as:
\begin{equation}
    \mathcal{L}^{CD} = \sum_{G_j^t \in \mathcal{B}^t} \frac{-1}{|\Pi(j)|} \sum_{i \in \Pi(j)} \log \frac{\exp \left(\bm{z}_j^t \ast \bm{z}_i^s / \tau\right)}{\sum_{G_{i'}^s\in \mathcal{B}^s} \exp \left(\bm{z}_j^t \ast \bm{z}_{i'}^s / \tau\right)}.
\label{cross-domain-loss}
\end{equation}

Our cross-domain contrastive learning objective has two benefits. On the one hand, given that the numerator of each term penalizes the distances between source samples and target samples with the same semantics, our loss contributes to generating domain-invariant graph representations. On the other hand, due to the promising results achieved by contrastive learning~\cite{you2020graph,li2020prototypical,khosla2020supervised,huang2021model}, comparing positive pairs with negative pairs can aid in developing discriminative graph representations for effective graph classification under label scarcity. We also construct the contrastive learning objective in the other branch and sum them to get the final loss. 
%\noindent\textbf{Theoretical Analysis.} 
In addition, we demonstrate that our cross-domain contrastive learning can be interpreted as maximizing the log-likelihood on target data using an Expectation Maximization (EM) algorithm. Compared with previous contrastive learning work~\cite{he2020momentum,chen2020simple}, our model utilizes a coupled framework which includes cross-module contrastive learning and cross-domain contrastive learning, which follow the paradigm of the EM algorithm while contrastive learning on images usually utilizes the InfoNCE loss to maximize the consistency.

\begin{proposition}
The cross-domain contrastive framework follows the EM algorithm. 
\label{prop}
\end{proposition}

The proof of Proposition \ref{prop} can be found in Appendix \ref{theory}.

\begin{algorithm}[t]
\caption{Learning Algorithm of \method{}}
\label{alg}
\begin{flushleft}
\textbf{Input:} Source data $\mathcal{D}^s$; Target data $\mathcal{D}^t$. \\
\textbf{Output}: GCN parameters $\theta$, HGKN parameters ${\phi}$, Classifier parameters $\eta$.  \\
\end{flushleft}
\begin{algorithmic}[1] 
\STATE Initialize model parameters.
\WHILE{not convergence}
    \STATE Sample mini-batches $\mathcal{B}^s$ and $\mathcal{B}^t$ from source and target data, respectively;
    \STATE Forward propagation $\mathcal{B}^s$ and $\mathcal{B}^t$ through two branches;
    \STATE Calculate the loss function in Eq. \ref{eq:final_loss};
    \STATE Update model parameters through back propagation;
\ENDWHILE
\end{algorithmic}
\end{algorithm}

\subsection{Summarization}
% To produce label distributions of test data, we utilize enhanced graph representations from the first branch to predict the label via a fully-connected layer. The reason for selecting the first branch is that the single GCN is more efficient than a single HGKN during evaluation. 

% In a nutshell, we combine two contrastive learning objectives with a supervised objective. The overall loss objective is written as:
% \begin{equation}\label{eq:final_loss}
%     \mathcal{L} = \mathcal{L}^{CB} +  \mathcal{L}^{CD} + \mathcal{L}^{S},
% \end{equation}
% where $\mathcal{L}^S=\frac{1}{|\mathcal{B}^s|}\sum_{G_i^s\in \mathcal{B}^s} H(\hat{y}_i^s, y_i^s)$, $H(\cdot, \cdot)$ denotes the cross entropy loss and $\hat{y}_i^s$ denotes the output of the classifier. 

% In \method{}, we adopt the standard cross entropy $H(\cdot, \cdot)$ to formulate the supervised objective as follows:
% \begin{equation}
%     \mathcal{L}^S=\frac{1}{|\mathcal{B}^s|}\sum_{G_i^s\in \mathcal{B}^s} H(\hat{y}_i^s, y_i^s),
% \end{equation}
% where $\hat{y}_i^s$ denotes the output of the classifier. 

To produce label distributions of test data, we utilize enhanced graph representations from the first branch to predict the label via a fully-connected layer. The reason for selecting the first branch is that the single GCN is more efficient than a single HGKN during evaluation. In \method{}, we adopt the standard cross entropy $H(\cdot, \cdot)$ to formulate the supervised objective as follows:
\begin{equation}
    \mathcal{L}^S=\frac{1}{|\mathcal{B}^s|}\sum_{G_i^s\in \mathcal{B}^s} H(\hat{y}_i^s, y_i^s),
\end{equation}
where $\hat{y}_i^s$ denotes the output of the classifier. 
In a nutshell, we combine two contrastive learning objectives with a supervised objective. The overall loss objective is:
\begin{equation}\label{eq:final_loss}
    \mathcal{L} = \mathcal{L}^{CB} +  \mathcal{L}^{CD} + \mathcal{L}^{S}.
\end{equation}

The algorithmic overview of \method{} is depicted in Algorithm \ref{alg}. The computing complexity of the proposed \method{} primarily relies on two networks. For a given graph $G$, $||A||_0$ denotes the number of nonzeros in the adjacency matrix. $d$ is the feature dimension. $L$ and $K$ denote the layer number of GCN and HGKN, respectively. $|V|$ is the number of nodes. $M$ denotes the number of filter graphs. 
The graph convolutional network takes $\mathcal{O}(L||A||_0d+L|V|d^2)$ computational time while the graph kernel module takes $\mathcal{O}(K|V|M)$ for each graph. 
As a result, the complexity of our \method{} is proportional to both $|V|$ and $||A||_0$.

\section{Experiments}

\begin{table*}[ht]
\footnotesize
\centering
\tabcolsep=0.8pt
\caption{The classification results (in \%) on Mutagenicity (source$\rightarrow$target).}
\resizebox{\textwidth}{!}{
\begin{tabular}{lccccccccccccc}
\toprule
{\bf Methods} &M0$\rightarrow$M1 &M1$\rightarrow$M0 &M0$\rightarrow$M2 &M2$\rightarrow$M0 &M0$\rightarrow$M3 &M3$\rightarrow$M0 &M1$\rightarrow$M2 &M2$\rightarrow$M1 &M1$\rightarrow$M3 &M3$\rightarrow$M1 &M2$\rightarrow$M3 &M3$\rightarrow$M2 &Avg.\\
\midrule
GCN &71.1 &70.4 &62.7 &69.0 &57.7 &59.6 &68.8 &74.2 &53.6 &63.3 &65.8 &74.5 &65.9\\
WL subtree  &74.9 &74.8 &67.3 &69.9 &57.8 &57.9 &73.7 &80.2 &60.0 &57.9 &70.2 &73.1 &68.1\\
CDAN &73.8 &74.1 &68.9 &71.4 &57.9 &59.6 &70.0 &74.1 &60.4 &67.1 &59.2 &63.6 &66.7\\
ToAlign &74.0 &72.7 &69.1 &65.2 &54.7 &73.1 &71.7 &77.2 &58.7 &73.1 &61.5 &62.2 &67.8\\
MetaAlign &66.7 &51.4 &57.0 &51.4 &46.4 &51.4 &57.0 &66.7 &46.4 &66.7 &46.4 &57.0 &55.4\\
GIN &72.3 &68.5 &64.1 &72.1 &56.6 &61.1 &67.4 &74.4 &55.9 &67.3 &62.8 &73.0 &66.3\\
CIN &66.8 &69.4 &66.8 &60.5 &53.5 &54.2 &57.8 &69.8 &55.3 &74.0 &58.9 &59.5 &62.2\\
GMT &73.6 &75.8 &65.6 &73.0 &56.7 &54.4 &72.8 &77.8 &62.0 &50.6 &64.0 &63.3 &65.8\\
% DEAL &76.3 &72.6 &69.8 &73.3  &58.3 &71.2 &\textbf{77.9} &80.8 &64.1 &74.1 &\textbf{70.6} &74.9 &72.0\\
DUA &70.2 &56.5 &64.0 &63.7 &53.6 &68.5 &57.7 &76.0 &65.1 &59.8 &57.9 &67.7 &63.4 \\
\midrule 

\method{}   &\textbf{77.7} &\textbf{76.6}  &\textbf{73.3} &\textbf{74.5} &\textbf{66.6} &\textbf{74.3} &77.3 &\textbf{80.8} &\textbf{67.4} &\textbf{74.1} &68.9 &\textbf{77.5} &\textbf{74.1}\\
\bottomrule
\end{tabular}}
\label{tab::results}
\vspace{-0.4cm}
\end{table*}

\begin{table*}[ht]
\footnotesize
\centering
\tabcolsep=3pt
\caption{The classification results (in \%) on Tox21 (source$\rightarrow$target).}
    \resizebox{\textwidth}{!}{
\begin{tabular}{lccccccccccccc}
\toprule
{\bf Methods} &T0$\rightarrow$T1 &T1$\rightarrow$T0 &T0$\rightarrow$T2 &T2$\rightarrow$T0 &T0$\rightarrow$T3 &T3$\rightarrow$T0 &T1$\rightarrow$T2 &T2$\rightarrow$T1 &T1$\rightarrow$T3 &T3$\rightarrow$T1 &T2$\rightarrow$T3 &T3$\rightarrow$T2 &Avg.\\
\midrule
GCN &64.2 &50.3 &67.9 &50.4 &52.2 &53.8 &68.7 &61.9 &59.2 &51.4 &54.9 &76.3 &59.3\\
WL subtree  &65.3 &51.1 &69.6 &52.8 &53.1 &54.4 &71.8 &65.4 &60.3 &61.9 &57.4 &76.3 &61.6\\
CDAN &69.9 &55.2 &78.3 &56.0 &59.5 &56.6 &78.3 &\textbf{68.5} &61.7 &68.1 &61.0 &78.3 &66.0\\
ToAlign &68.2 &58.5 &78.4 &58.8 &58.5 &53.8 &\textbf{78.8} &67.1 &64.4 &68.8 &57.9 &78.4 &66.0\\
MetaAlign &65.7 &57.5 &78.0 &58.5 &\textbf{63.9} &52.2 &78.8 &67.1 &62.3 &67.5 &56.8 &78.4 &65.6\\
GIN &67.8 &51.0 &77.5 &54.3 &56.8 &54.5 &78.3 &63.7 &56.8 &53.3 &56.8 &77.1 &62.3\\
CIN &67.8 &50.3 &78.3 &54.5 &56.8 &54.5 &78.3 &67.8 &59.0 &67.8 &56.8 &78.3 &64.2\\
GMT &67.8 &50.0 &78.4 &50.1 &56.8 &50.7 &78.3 &67.8 &56.8 &67.8 &56.4 &78.1 &63.3\\
% DEAL &73.9 &59.4 &68.2 &58.2 &53.8 &58.7 &69.2 &66.2 &63.5 &67.4 &61.1 &77.8 &64.8\\
DUA &60.6 &51.3 &70.7 &52.5 &53.6 &49.3 &71.1 &67.2 &53.6 &59.0 &58.6 &74.3 &60.2 \\
\midrule 
\method{}   &\textbf{69.9} &\textbf{59.8} &\textbf{78.8}  &\textbf{59.0} &62.3 &\textbf{59.0} &78.4 &66.8 &\textbf{65.0} &\textbf{68.8} &\textbf{61.2} &\textbf{78.4} &\textbf{67.3}\\
\bottomrule
\end{tabular}}
\label{tab::results1}
\vspace{-0.4cm}
\end{table*}

\begin{table}[ht]
\footnotesize
\centering
\tabcolsep=1pt
\caption{The classification results (in \%) on PROTEINS, COX2, and BZR (source$\rightarrow$target).}
\begin{tabular}{lccccccc}
\toprule
{\bf Methods} &P$\rightarrow$D &D$\rightarrow$P &C$\rightarrow$CM &CM$\rightarrow$C &B$\rightarrow$BM &BM$\rightarrow$B &Avg.\\
\midrule
GCN &58.7  &59.6  &51.1 &78.2 &51.3 &71.2 &61.7\\
WL subtree  &72.9 &41.1 &48.8 &78.2 &51.3 &78.8 &61.9\\
CDAN &59.7 &64.5 &59.4 &78.2 &57.2 &78.8 &66.3\\
ToAlign &62.6 &64.7 &51.2 &78.2 &58.4 &78.7 &65.7\\
MetaAlign &60.3 &64.7 &51.0 &77.5 &53.6 &78.5 &64.3\\
GIN &61.3 &56.8 &51.2 &78.2 &48.7 &78.8 &62.5\\
CIN &62.1 &59.7 &57.4 &61.5 &54.2 &72.6 &61.3\\
GMT &62.7 &59.6 &51.2 &72.2 &52.8 &71.3 &61.6\\
% DEAL  &\textbf{76.2} &63.6 &\textbf{62.0} &78.2  &58.5 &78.8 &69.6 \\
DUA &61.3 &56.9 &51.3 &69.5 &56.4 &70.2 &60.9\\
\midrule 
% \method{}-GCN &73.0 &56.9 &60.7 &\textbf{79.2} &52.6 &\textbf{80.7} \\
% \method{}-SAGE \\
 \method{}  &74.6 &\textbf{67.0} &61.1 &\textbf{79.0} &\textbf{62.7} &\textbf{78.8} &\textbf{70.5}\\
\bottomrule
\end{tabular}
\label{tab::results2}
\vspace{-0.4cm}
\end{table}

% In this section, we conduct extensive experiments on various datasets to verify the effectiveness of the proposed \method{}. 
% We aim to answer the following questions:
% (1) How does the proposed \method{} perform compared with the state-of-the-art baseline methods for unsupervised domain adaptive graph classification? 
% (2) How do different GCN architectures and graph kernels affect the performance of the proposed \method{}?
% (3) What is the contribution of each submodule to the final result in the proposed \method{}.
% (4) How do the hyper-parameters affect the performance of the proposed \method{}?

\begin{table*}[ht]
\footnotesize
\centering
\tabcolsep=0.5pt
\caption{The results of ablation studies on Mutagenicity (source$\rightarrow$target).}\label{tab::ablation}
\resizebox{\textwidth}{!}{
\begin{tabular}{lccccccccccccc}
\toprule
{\bf Methods} &M0$\rightarrow$M1 &M1$\rightarrow$M0 &M0$\rightarrow$M2 &M2$\rightarrow$M0 &M0$\rightarrow$M3 &M3$\rightarrow$M0 &M1$\rightarrow$M2 &M2$\rightarrow$M1 &M1$\rightarrow$M3 &M3$\rightarrow$M1 &M2$\rightarrow$M3 &M3$\rightarrow$M2 &Avg.\\
\midrule
\method{}/CB &\textbf{78.0} &72.2 &63.8 &71.4 &62.1 &69.1 &76.0 &77.3 &64.1 &74.8 &65.4 &76.3 &70.9\\
\method{}/CD &74.0 &72.7 &65.8 &65.9 &58.6 &69.4 &71.5 &78.3 &66.6 &74.9 &66.2 &75.6 &70.0\\
\method{}-GIN &74.9 &73.6 &63.6 &70.7 &58.3 &70.0 &76.6 &78.5 &66.5 &\textbf{75.1} &66.7 &77.5 &71.0\\
\method{}-HGKN &74.9 &73.0 &64.5 &70.5 &63.0 &70.6 &76.3 &78.0 &66.1 &74.7 &66.4 &77.3 &71.3\\
\midrule
\method{}   &77.7 &\textbf{76.6}  &{73.3} &\textbf{74.5} &\textbf{66.6} &{74.3} &{77.3} &{80.8} &\textbf{67.4} &74.1 &{68.9} &\textbf{77.5} &\textbf{74.1}\\
\midrule
\method{}-NP &77.2 &76.0 &\textbf{74.1} &74.2 &65.5 &\textbf{74.6} &\textbf{77.7} &\textbf{81.0} &67.1 &{74.7} &\textbf{69.2} &76.1 &74.0\\
\bottomrule 
\end{tabular}
}
\end{table*}

\subsection{Experimental Settings}

\noindent\textbf{Datasets.}
We perform experiments on various real-world benchmark datasets (i.e., Mutagenicity, Tox21, PROTEINS, DD, BZR and COX2) from TUDataset \cite{Morris2020} in the setting of unsupervised domain adaptation. For convenience, P, D, C, CM, B, and BM are short for PROTEINS, DD, COX2, COX2\_MD, BZR, and BZR\_MD, respectively. Their details are introduced as follows:
\begin{itemize}[leftmargin=*]
    \item \textbf{Mutagenicity}~\cite{kazius2005derivation}: Mutagenicity is a popular dataset that consists of 4337 molecular structures and their corresponding Ames test data.
    To distinguish the distribution of datasets, we divide it into four sub-datasets (i.e., M0, M1, M2 and M3) based on the edge density.
    \item \textbf{Tox21\footnote{https://tripod.nih.gov/tox21/challenge/data.jsp}}: The purpose of the Tox21 dataset is to assess the predictive ability of models in detecting compound interferences. We separate the dataset into four sub-datasets according to their interactions with `aromatase' and `HSE'. 
    \item \textbf{PROTEINS}: PROTEINS~\cite{dobson2003distinguishing} and DD~\cite{shervashidze2011weisfeiler} are investigated here and each label indicates if a protein is a non-enzyme or not. 
    The presentation of each protein is in the form of a graph, where the amino acids serve as nodes and edges exist when the distance between two nodes is less than 6 Angstroms.
    \item \textbf{COX2}: We investigate datasets COX2 and COX2\_MD~\cite{sutherland2003spline} which consists of 467 and 303 cyclooxygenase-2 inhibitors. Every sample characterizes a chemical compound, with edges determined by distance and vertex features representing atom types.
    \item \textbf{BZR}: We explore the BZR and BZR\_MD \cite{sutherland2003spline} datasets, which contains ligands for the benzodiazepine receptor.
\end{itemize}
\vspace{-0.4cm}

\noindent\textbf{Baselines.}
To increase persuasiveness, we compare the proposed \method{} with a large number of state-of-the-art methods, including one graph kernel approach ({WL subtree}~\cite{shervashidze2011weisfeiler}), four graph neural network methods ({GCN}~\cite{kipf2017semi}, {GIN}~\cite{xu2019powerful}, {CIN}~\cite{bodnar2021weisfeiler} and {GMT}~\cite{BaekKH21}), and four recent domain adaptation methods ({CDAN}~\cite{long2018conditional}, {ToAlign}~\cite{NeurIPS2021_731c83db}, {MetaAlign}~\cite{wei2021metaalign} and DUA~\cite{mirza2022dua}). Their detailed introduction is elaborated in Appendix \ref{baseline}.

\noindent\textbf{Implementation Details.}
\label{setting}
We employ a two-layer GIN~\cite{xu2019powerful} in the GCN branch and a two-layer network along with the WL kernel~\cite{shervashidze2011weisfeiler} in the HGKN branch. We use the Adam as the default optimizer and set the learning rate to $10^{-4}$. The embedding dimension of hidden layers and batch size are both set to 64. 
The pooling ratio $\rho$ and the number of filter graphs $M$ are set to $0.4$ and $15$, respectively.
For the sake of fairness, we employ the same GCN as the graph encoder. In the graph classification baselines in all domain adaption baselines. We utilize all the labeled source samples to train the model and evaluate unlabeled samples when it comes to graph classification methods as in \cite{wu2020unsupervised}. 
We initialize the parameters of all the compared methods as in their corresponding papers and fine-tune them to achieve the best performance.
% The parameters in compared baselines are initialized according to the corresponding papers and are tuned for the best performance afterward. 

% \begin{figure}[t]
%   \centering
%   \includegraphics[scale=0.64]{img/gnn.pdf}
%   \caption{The performance with different GCN architectures and graph kernels on Mutagenicity.}
%   \label{fig:GKN}
% \end{figure}

\subsection{Performance Comparison}
Table \ref{tab::results}, \ref{tab::results1} and \ref{tab::results2} show the performance of graph classification in various settings of unsupervised domain adaptation. From the results, we observe that: 
% \begin{itemize}[leftmargin=*]
1) The domain adaptation methods generally perform better than graph kernel and GNN methods in Table \ref{tab::results1} and \ref{tab::results2}, indicating that current graph classification models are short of the capacity of transfer learning. Thus, it is essential to design an effective domain adaptive framework for graph classification.
2) Although the domain adaptation methods obtain competitive results on simple transfer tasks in Table \ref{tab::results2}, it is difficult for them to make great progress on hard transfer tasks in comparison with GIN (Table \ref{tab::results}). We attribute the reason to the immense difficulty of acquiring graph representations. Therefore, it is unwise to directly apply current domain adaptation approaches to GCNs.
3) Our \method{} outperforms all the baselines in most cases. In particular, the average improvement of CoCo ranges is 8.81\% on Mutagenicity. We attribute the performance gain to two key points: (i) Contrastive learning across coupled branches (i.e., GCN and HGKN) improves the representation ability of graphs under label scarcity. (ii) Cross-domain contrastive learning contributes to effective domain alignment from an EM perspective.
% \end{itemize}
% The performance gain comes from two aspects: (1) Introduction of adversarial perturbation. \method{} adds perturbations on source graphs, which achieves effective domain alignment with the exploration of underlying semantic priors. 
% (4) Introduction of pseudo-label distilling. To mitigate the effects of noisy pseudo-labels for target data, we explore different feature spaces from graph neural networks to distill consistent predictions for reliable semantic learning. 

\subsection{Effect of Different GNNs and Graph Kernels}
\label{effect}
\begin{figure}[t]
  \centering
  \includegraphics[scale=0.5]{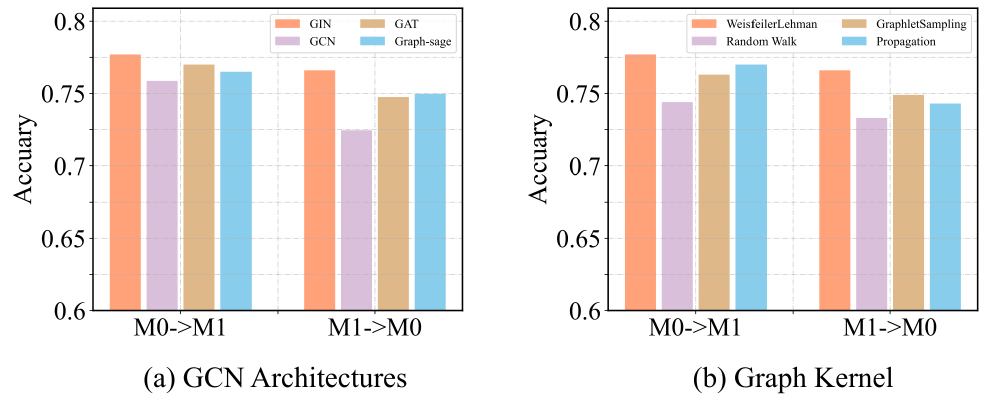}
    \vspace{-0.5cm}
  \caption{The performance with different GCN architectures and graph kernels on Mutagenicity.}
  \label{fig:GKN}
\end{figure}

To investigate the flexibility of \method{}, we replace the GIN in our implementation with different GNN methods (i.e., GCN~\cite{kipf2017semi}, GAT~\cite{velickovic2018graph} and Graphsage~\cite{hamilton2017inductive}) and the WL kernel with different graph kernels (i.e., Graph Sampling~\cite{leskovec2006sampling}, Random Walk~\cite{kalofolias2021susan} and Propagation~\cite{neumann2016propagation}).
Figure~\ref{fig:GKN} shows the performance of different GNNs and graph kernels on two representative datasets, and we have similar observations on other datasets. From the results, we have the following observation, by comparing with other GCN architectures and graph kernels, GIN and WL kernels achieve the best performance in most of the cases and the reason can be attributed to the powerful representation capability of GIN and WL kernel. This also justifies the motivation why we select WL kernel to improve the performance in our task of domain adaptation. 

\begin{figure}[t]
  \centering
  \includegraphics[scale=0.63]{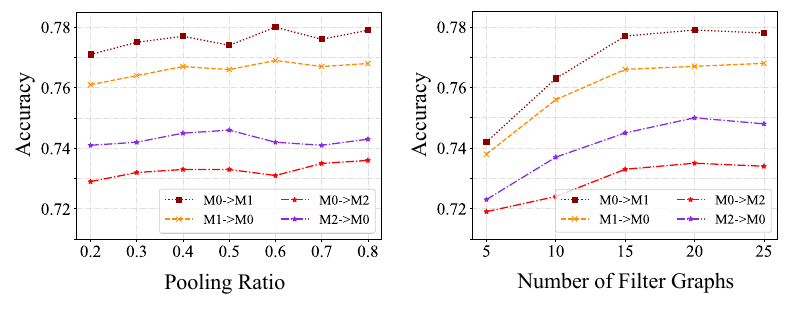}
    \vspace{-0.5cm}
  \caption{Sensitivity analysis on Mutagenicity. We select four transfer learning tasks here. }
  \label{fig:hyperparameter}
\end{figure}

\subsection{Ablation Study}

To evaluate the impact of each component in our \method{}, we introduce several model variants as follows: 
% \begin{itemize}[leftmargin=*]
1) \textbf{\method{}}/\textbf{CB}: It removes the cross-branch contrastive learning module;
2) \textbf{\method{}}/\textbf{CD}: It removes the cross-domain contrastive learning module; 
3) \textbf{\method{}-GIN}: It uses two distinct GINs to generate coupled graph representations; 
4) \textbf{\method{}-HGKN}: It uses two distinct HGKN to generate coupled graph representations.
5) \textbf{\method{}-\textbf{NP}}: It utilizes the non-parametric classifier instead of the MLP classifier for target domain prediction.
% \end{itemize}

We compare the performance on Mutagenicity and the results are shown in Table~\ref{tab::ablation}. From Table~\ref{tab::ablation}, we have the following observations:
% \begin{itemize}[leftmargin=*]
1) \method{}/CB exhibits inferior performance compared with the \method{} in most settings, validating that comparing complementary information extracted by GCN and HGKN is capable of improving graph representation learning. 
2) \method{}/CD obtain the worst performance among four model variants, we attribute the reason to that the cross-domain contrastive learning module aligns the source and target graph representations effectively, which tackles serious domain discrepancy in the graph space. This challenge is the most crucial in our task.  
3) \method{}-GIN and \method{}-HGKN achieves approximate results but performs worse than \method{}, which demonstrates that merely ensemble learning in the implicit (explicit) manner cannot largely enhance graph representation learning under label scarcity.
4) The average performance of \method{} and \method{}-NP is quite similar. The reason is that after sufficient training, the performance of the non-parametric classifier and the MLP classifier is similar. However, the non-parametric classifier demands a substantial amount of labeled source domain data, resulting in significantly lower efficiency compared to the MLP classifier. Consequently, we recommend employing the MLP classifier for target domain prediction.
% \end{itemize}
% slightly outperforms \method{}/T buts obtains much worse performance compared with our full model. The potential reason could be that the pseudo-labels obtained from the classifier are noisy and over-confident, which can merely bring in limited performance improvement. In comparison, our pseudo-label distilling module can generate accurate pseudo-labels.
% (4) \method{}-F and \method{}-H both perform worse than the full model, which implies the importance of adding composite perturbations to the data. 

\subsection{Sensitivity Analysis}
In this part, we investigate the sensitivity of the proposed \method{} to the hyper-parameters (i.e., pooling ratio $\rho$ and number of filter graphs $M$) that affect the performance of \method{}. $\rho$ determines the number of kept nodes of pooled graphs, and $M$ denotes the number of filters. Figure \ref{fig:hyperparameter} plots the results on Mutagenicity. 
We first vary $\rho$ in the set from 0.2 to 0.8 with other parameters fixed. The performance generally shows an increasing trend at the beginning and stabilizes as the ratio of $\rho$ increases.
% It can be seen that with the ratio of $\rho$ rises, the performance tends to increase at the beginning and then remains stable in most of the cases. 
We attribute the reason to the small $\rho$ keeping fewer nodes which may miss some important information. On the contrary, the larger $\rho$ retains more node information, but the complexity of the model is greatly increased. Considering the trade-off between model performance and complexity, we set $\rho$ to 0.4 as default.
In addition, We vary $M$ in \{5, 10, 15, 20, 25\} while fixing other parameters. It can be found that with the increasing of $M$, the \method{} achieves better performance when the value is small. The potential reason is that increasing the number of filter graphs would help to extract high-order structure information.
Nevertheless, when $M$ is too large (i.e., $M>15$), the improvement in performance is not obvious. Thus, to balance the performance with the computational complexity, we set $M$ to 15.
% \end{itemize}

% However, too large of $M$ may hurt the model performance, which demonstrates that too strong perturbations could harm the discrimination information of source data and thus lead to ineffective domain adaptive graph classification.
\section{Conclusion}
This paper addresses the practical problem of unsupervised graph classification and introduces an effective method named \method{} is proposed. At a high level, \method{} is featured by two branches, i.e., a graph convolutional network branch and a hierarchical graph kernel network branch, which explores graph topological information in implicit and explicit manners, respectively. Then, we integrate the two branches into a multi-view contrastive learning framework where cross-branch contrastive learning aims to generate discriminative graph representation with full semantics whereas cross-domain contrastive learning strives to minimize domain discrepancy. Extensive experiments on diverse datasets validate the efficacy of proposed \method{} compared with various competing methods. In future works, we will extend our proposed \method{} to address more complicated tasks, including domain generalization, multi-source domain adaptation and source-free domain adaptation.

% \verb|\end{document}|

\bibliography{example_paper}
\bibliographystyle{icml2023}
\newpage
\appendix
\onecolumn
\appendix
\section{An Expectation-Maximization Perspective}
\label{theory}
Maximum Likelihood (ML) has been widely utilized in various unsupervised machine learning problems~\cite{sharma2016hierarchical,li2020prototypical,gresele2020relative}. In our settings, it aims to find the optimal weights of the graph encoder $\theta^*$ to maximize the log-likelihood of target graphs within a mini-batch. In formulation, the log-likelihood function is written as:
\begin{equation}
    \ell_{likelihood}(\theta) = \sum_{G_j^t\in \mathcal{B}^t}\log p(G_j^t;\theta).
\end{equation}
To make use of abundant labeled source graphs, the unlabeled sample $G_j^t \in \mathcal{B}^t$ are compared with source graphs in a mini-batch, denoted as $\{G_i^s\}_{i=1}^{B^s}$ for clearness, and we have:
\begin{equation}\label{eq15}
    \theta^*=\mathop{\arg\max}\limits_{\theta}\sum_{G_j^t\in \mathcal{B}^t}\log \sum_{i=1}^{B^s} p(G_j^t,G_i^s;\theta).
\end{equation}
However, directly optimizing the function would be difficult. To tackle this, we introduce a surrogate function $Q(G_i^s)$ $(\sum_{i=1}^{B^s}Q(G_i^s)=1)$ to estimate the lower-bound of Eq. \ref{eq15}. By applying Jensen's inequality, we have:
\begin{equation}
\small
\begin{aligned}
    \sum_{G_j^t\in \mathcal{B}^t}\log \sum_{i=1}^{B^s} p(G_j^t,G_i^s;\theta)&=\sum_{G_j^t\in \mathcal{B}^t}\log \sum_{i=1}^{B^s} Q(G_i^s)\frac{p(G_j^t,G_i^s;\theta)}{Q(G_i^s)}\\
    &\ge \sum_{G_j^t\in \mathcal{B}^t}\sum_{i=1}^{B^s} Q(G_i^s)\log  \frac{p(G_j^t,G_i^s;\theta)}{Q(G_i^s)}.
\end{aligned}
\label{bond}
\end{equation}
The equality holds when $Q(G_i^s)/ p(G_i^s,G_j^t;\theta)$ is a constant, which results in the following formulation:
\begin{equation}
    \frac{Q(G_i^s)}{p(G_i^s,G_j^t;\theta)}=\frac{\sum_{i=1}^{B^s} Q(G_i^s) }{\sum_{i=1}^{B^s} p(G_i^s,G_j^t;\theta)} = \frac{1}{p(G_j^t;\theta)}.
\end{equation}
Hence, we have $Q(G_i^s)=p(G_i^s;G_j^t,\theta)$.

It is worth noting that $-\sum_{G_j^t\in\mathcal{B}^t}\sum_{i=1}^{B^s} Q(G_i^s) \log Q(G_i^s)$ does not influence the optimization of $\theta$. Therefore, we rewrite the objective function as follows:
\begin{equation}
    \ell=\sum_{G_j^t\in \mathcal{B}^t}\sum_{i=1}^{B^s} p(G_i^s;G_j^t,\theta)\log  p(G_j^t,G_i^s;\theta),
\label{all}
\end{equation}
Then, we utilize an EM algorithm to maximize Eq. \ref{all}. 

\noindent\textbf{\textit{E step.}} 
This step aims to infer the posterior probability $p(G_i^s;G_j^t,\theta)$. To begin with, we calculate the pseudo-label of the target data $G_j^t$ using $g_{{\theta}}$ where ${\theta}$ denotes the current model parameters. Treating all source samples with the same label equally, we have $p(G_i^s;G_j^t,\theta)=\frac{1}{|\Pi(j)|}\mathbbm{1}(G_j^t,G_i^s)$ where $\mathbbm{1}(G_j^t,G_i^s)=1$ if they has the same label and $|\Pi(j)|=\sum_{i=1}^{B^s} \mathbbm{1}(G_j^t,G_i^s) $ denotes the number of source samples with the same label. 

\noindent\textbf{\textit{M step.}}
This step aims to maximize the lower-bound of Eq. \ref{all} based on E-step. In particular, we have:
\begin{equation}
\begin{aligned}
\ell &=\sum_{G_j^t\in \mathcal{B}^t}\sum_{i=1}^{B^s} p(G_i^s;G_j^t,\theta)\log  p(G_j^t,G_i^s;\theta)\\
    &=\sum_{G_j^t\in \mathcal{B}^t}\sum_{i=1}^{B^s} \frac{1}{|\Pi(j)|}\mathbbm{1}(G_j^t,G_i^s)\log  p(G_j^t,G_i^s;\theta).
    \end{aligned}
    \label{Mstep}
\end{equation}
When presuming the prior obeys a uniform distribution and employing an isotropic Gaussian to characterize the distribution of each sample in the embedding space, we have:
\begin{equation}
\small
\begin{aligned}
p(G_j^t,G_i^s;\theta&)=p(G_j^t;G_i^s,\theta)p(G_i^s;\theta)=\frac{1}{{B^s}}\cdot p(G_j^t;G_i^s,\theta),\\
    p(G_j^t;G_i^s,
    \theta)&=\exp (\frac{-(\bm{z}_j^t-\bm{z}_i^s)^2}{2\sigma_i^2})/\sum_{i'=1}^{B^s}\exp(\frac{-(\bm{z}_j^t-\bm{z}^s_{i'})^2}{2\sigma_{i'}^2}),
    \label{poster}
\end{aligned}
\end{equation}
where $\bm{z}_j^t=g_\phi(G_j^t)$ and $\bm{z}_i^s = g_\phi(G_i^s)$. $\sigma_{i}^2$ denotes the variance of the Gaussian distribution around $\bm{z}_i^s$, which is assumed the same for all source samples. Hence, we set $\tau = \sigma_{i}^2$. 
Assuming $l_2$-normalized $\bm{z}_j^t$ and $\bm{z}_i^s$, we obtain $(\bm{z}_j^t-\bm{z}_i^s)^2=2-2\bm{z}_j^t \ast \bm{z}_i^s$. 
By incorporating the Eq \ref{Mstep} and Eq. \ref{poster} into Eq. \ref{eq15}, we have:
\begin{equation}
    \small
    \theta=\mathop{\arg\max}\limits_{\theta}\sum_{G_j^t\in\mathcal{B}^t}\frac{1}{|\Pi(j)|}\sum_{i=1}^{B^s} \mathbbm{1}(G_j^t,G_i^s) \log \frac{\exp(\bm{z}_j^t\ast{\bm{z}_i^s}/\tau)}{\sum_{i'=1}^{B^s} \exp(\bm{z}_j^t\ast{\bm{z}^s_{i'}}/\tau)}
\label{org}
\end{equation}
which is equivalent to our cross-domain contrastive learning objective in Eq. \ref{cross-domain-loss}.

% \begin{algorithm}[t]
% \caption{Learning Algorithm of \method{}}
% \label{alg}
% \begin{flushleft}
% \textbf{Input:} Source data $\mathcal{D}^s$; Target data $\mathcal{D}^t$. \\
% \textbf{Output}: GCN parameters $\theta$, HGKN parameters ${\phi}$, Classifier parameters $\eta$.  \\
% \end{flushleft}
% \begin{algorithmic}[1] 
% \STATE Initialize model parameters.
% \WHILE{not convergence}
%     \STATE Sample mini-batches $\mathcal{B}^s$ and $\mathcal{B}^t$ from source and target data, respectively;
%     \STATE Forward propagation $\mathcal{B}^s$ and $\mathcal{B}^t$ through two branches;
%     \STATE Calculate the loss function in Eq. \ref{eq:final_loss};
%     \STATE Update model parameters through back propagation;
% \ENDWHILE
% \end{algorithmic}
% \end{algorithm}
\begin{figure}[t]
  \centering
  \includegraphics[scale=0.64]{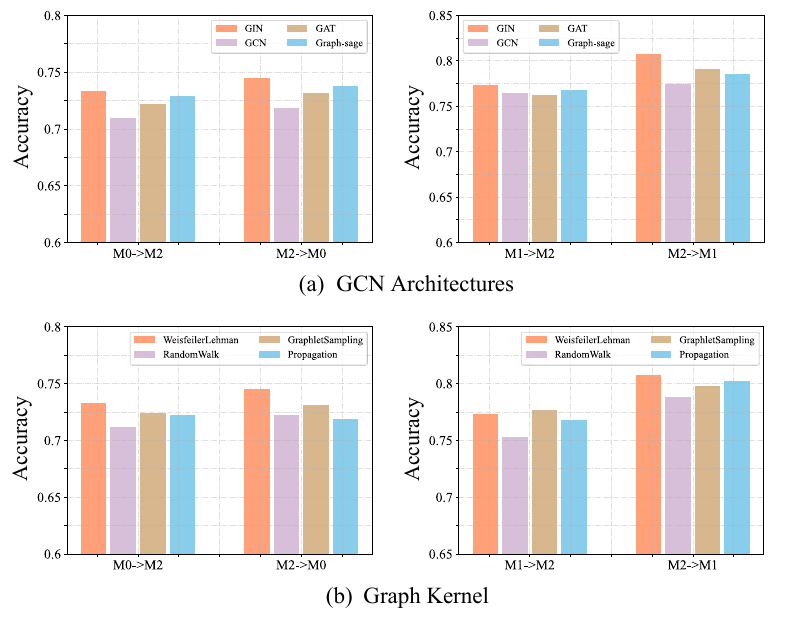}
  \caption{The performance with different GCN architectures and graph kernels on Mutagenicity.}
%   \vspace{-0.4cm}
  \label{fig:more}
\end{figure}

\section{Introduction of Baselines}\label{baseline}
\begin{itemize}[leftmargin=*]
    \item \textbf{WL subtree}~\cite{shervashidze2011weisfeiler}: WL subtree is a kernel method, which measures the similarity between graphs with the defined kernel function.
    % It extracts features based on the WL test, which maps each graph to a sequence of new graphs with node attributes capturing label and topological information.
    \item \textbf{GCN}~\cite{kipf2017semi}: The main idea of GCN is to combine neighborhood information to update each central node, so as to obtain a representation vector in an iterative fashion.
    % It is a popular neural network that adopts a first-order approximation of spectral graph convolution to effectively encode structural information and features of nodes. 
    \item \textbf{GIN}~\cite{xu2019powerful}: GIN is a well-known message passing neural networks with improved expressive capability.
    % and generalizes the WL test to demonstrate the improved representation ability of exploring diverse graph topology.
    \item \textbf{CIN}~\cite{bodnar2021weisfeiler}: CIN extends the theoretical results on Simplicial Complexes to regular Cell Complexes, which achieves a better performance. 
    % It combines theoretical findings on Simplicial Complexes with graph neural networks and employs a hierarchical message passing mechanism to enhance the model capacity.
    \item \textbf{GMT}~\cite{BaekKH21}: GMT is a multi-head attention-based model, which captures interactions between nodes according to their structural dependencies.
    % It is comprised of an attention-based layer that attempts to distinguish the structural relationships between vertices in the graph.
    \item \textbf{CDAN}~\cite{long2018conditional}: CDAN utilizes a adversarial adaptation framework conditioned on discriminative information extracted from the predictions of the classifier.
    % It is a representative unsupervised domain adaption model which conditions the adversarial adaptation module on discriminative knowledge embedded in the classifier predictions.
    \item \textbf{ToAlign}~\cite{NeurIPS2021_731c83db}: ToAlign aims to align the domain by performing feature decomposition and the prior knowledge including classification task itself.
    % It performs feature decomposition and alignment of the source and target domain under the guidance of the prior knowledge induced from the classification task.
    \item \textbf{MetaAlign}~\cite{wei2021metaalign}: MetaAlign separates the domain alignment and the classification objectives as two individual tasks, i.e., meta-train and  meta-test, and uses a meta-optimization method to optimize these two tasks.
     \item\textbf{DUA}~\cite{mirza2022dua}: DUA proposes an effective normalization technique for domain adaptation. 
    % It utilizes a meta-optimization-based strategy that treats the domain alignment objective and the classification objective as the meta-train and meta-test tasks respectively.
\end{itemize}

% \section{Algorithm}

% The overall framework of \method{} is illustrated in Algorithm \ref{alg}. The computing complexity of our \method{} mainly depends on two networks. Given a graph $G$, $||A||_0$ is the number of nonzeros in the adjacency matrix. $d$ is the feature dimension. $L$ and $K$ is the layer number of GCN and HGKN, respectively. $|V|$ is the number of nodes. $M$ denotes the number of filter graphs. 
% The graph convolutional network takes $\mathcal{O}(L||A||_0d+L|V|d^2)$ computational time while the graph kernel module takes $\mathcal{O}(K|V|M)$ for each graph. 
% As a result, the complexity of our \method{} is linearly related to both $|V|$ and $||A||_0$.  

\section{Additional Experiments}

We conduct more experiments to evaluate the effect of different GCN architectures and Graph kernels. Figure \ref{fig:more} shows the performance of different GCN architectures and graph kernels on Mutagenicity. From Figure \ref{fig:more}, we have the similarity observations as described in Section \ref{effect}, validating the effectiveness of GIN and WL kernel once again. 
%%%%%%%%%%%%%%%%%%%%%%%%%%%%%%%%%%%%%%%%%%%%%%%%%%%%%%%%%%%%%%%%%%%%%%%%%%%%%%%
%%%%%%%%%%%%%%%%%%%%%%%%%%%%%%%%%%%%%%%%%%%%%%%%%%%%%%%%%%%%%%%%%%%%%%%%%%%%%%%

%%%%%%%%%%%%%%%%%%%%%%%%%%%%%%%%%%%%%%%%%%%%%%%%%%%%%%%%%%%%%%%%%%%%%%%%%%%%%%%
%%%%%%%%%%%%%%%%%%%%%%%%%%%%%%%%%%%%%%%%%%%%%%%%%%%%%%%%%%%%%%%%%%%%%%%%%%%%%%%
% APPENDIX
%%%%%%%%%%%%%%%%%%%%%%%%%%%%%%%%%%%%%%%%%%%%%%%%%%%%%%%%%%%%%%%%%%%%%%%%%%%%%%%
%%%%%%%%%%%%%%%%%%%%%%%%%%%%%%%%%%%%%%%%%%%%%%%%%%%%%%%%%%%%%%%%%%%%%%%%%%%%%%%
\newpage
\appendix
\onecolumn

% You can have as much text here as you want. The main body must be at most $8$ pages long.
% For the final version, one more page can be added.
% If you want, you can use an appendix like this one, even using the one-column format.
%%%%%%%%%%%%%%%%%%%%%%%%%%%%%%%%%%%%%%%%%%%%%%%%%%%%%%%%%%%%%%%%%%%%%%%%%%%%%%%
%%%%%%%%%%%%%%%%%%%%%%%%%%%%%%%%%%%%%%%%%%%%%%%%%%%%%%%%%%%%%%%%%%%%%%%%%%%%%%%

\end{document}